\documentclass[sigconf]{acmart}

\usepackage{booktabs} 

\setcopyright{rightsretained}

\usepackage{amsmath} 
\usepackage{amssymb}  
\usepackage[linesnumbered,ruled]{algorithm2e}
\usepackage{tikz}
\usepackage{graphicx}
\usepackage{wrapfig}
\usepackage{romannum}
\usepackage{textcomp}
\usepackage{hyperref}
\usepackage{subcaption}
\usepackage{diagbox}

\DeclareMathOperator*{\argmin}{arg\,min}

\acmConference[CoDS-COMAD]{Conference on Data Science}{Jan 2018}{Goa, India} 
\acmYear{2018}
\copyrightyear{2018}

\graphicspath{{images/}}

\begin{document}
\title{Bayesian Optimisation with Prior Reuse for Motion Planning in Robot Soccer}

\author{Abhinav Agarwalla}
\email{abhinavagarwalla@iitkgp.ac.in}
\orcid{1234-5678-9012}
\affiliation{%
  \institution{Indian Institute of Technology Kharagpur}
}

\author{Arnav Kumar Jain}
\email{arnavkj95@iitkgp.ac.in}
\affiliation{%
  \institution{Indian Institute of Technology Kharagpur}
}

\author{KV Manohar}
\email{kvmanohar22@iitkgp.ac.in}
\affiliation{%
  \institution{Indian Institute of Technology Kharagpur}
}

\author{Arpit Tarang Saxena}
\email{arpit.tarang@gmail.com}
\affiliation{%
  \institution{Indian Institute of Technology Kharagpur}
}

\author{Jayanta Mukhopadhyay}
\email{jay@cse.iitkgp.ac.in}
\affiliation{%
  \institution{Indian Institute of Technology Kharagpur}
}

\renewcommand{\shortauthors}{A. Agarwalla et al.}
\begin{abstract}
We integrate learning and motion planning for soccer playing differential drive robots using Bayesian optimisation. Trajectories generated using end-slope cubic B\'{e}zier splines are first optimised globally through Bayesian optimisation for a set of candidate points with obstacles. The optimised trajectories along with robot and obstacle positions and velocities are stored in a database. The closest planning situation is identified from the database using k-Nearest Neighbour approach. It is further optimised online through reuse of prior information from previously optimised trajectory. Our approach reduces computation time of trajectory optimisation considerably. Velocity profiling generates velocities consistent with robot kinodynamoic constraints, and avoids collision and slipping. Extensive testing is done on developed simulator, as well as on physical differential drive robots. Our method shows marked improvements in mitigating tracking error, and reducing traversal and computational time over competing techniques under the constraints of performing tasks in real time.
\end{abstract}

%
%



\maketitle


\section{Introduction}
The environment of a robot soccer game is highly dynamic and adversarial with fast moving different drive robots, and a competing opponent team. High-level complex manoeuvres like attacking and passing are performed through coordination of multiple robots making efficient planning even more challenging. Consider a case where a robot has to intercept a moving ball and shoot it to goal. The robot needs to reach the set interception point at the same time instant as the ball. High deviations from the generated trajectory would lead to missing the ball which is highly undesirable. Also, there is an opponent team whose bots are racing to be the first to reach the ball. This motivates the need for trajectory optimisation to intercept the ball in least possible time. Both trajectory traversal time and trajectory optimisation time contribute to interception time, and must be reduced. In this paper, we address this issue and determine time-efficient computation of trajectories for differentiable drive robots under real time constraints.

\begin{figure}[t]
\centering
\includegraphics[scale=0.5]{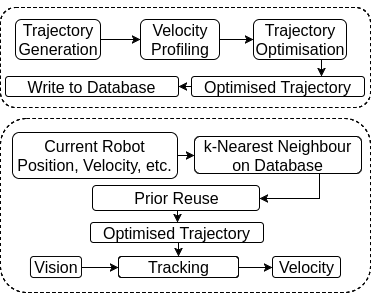}
\caption{Planning Overview: Initial trajectory generated is optimised using Bayesian optimisation for a combination of various starting and end points, and saved to a database. In the online optimisation, the closest prior is queried using k-Nearest Neighbour algorithm and trajectory is further optimised. Using the optimised trajectory and vision input, tracker calculates and sends robot velocities.}
\label{fig:overview}
\end{figure}

Previous works in robotics have addressed the problem of trajectory generation and tracking in a dynamic environment. These approaches were built on joining points by lines to avoid obstacles~\cite{borenstein1991vector,meng1992neural}, and later extended by interpolating the points using curves to make the path differentiable and smooth at the way-points~\cite{mahkovic1997smooth,saska2006transformed}. However, the path is not globally optimal with respect to time, and also may not be free of collision. Recent works like Dynamic Window~\cite{references:seder2007dynamic} and Potential Field~\cite{khatib1986real} are more focused on locally optimal trajectories to avoid collision with obstacles. Particle Swarm Optimization(PSO)~\cite{saska2006robot} have been used to optimise trajectories, but take a lot of time to converge. This makes them unsuitable for real time optimisation such as our use case. We adopt Bayesian optimisation for trajectory optimisation since it is a global optimisation technique requiring very few evaluations of objective function. Fig. \ref{fig:overview} gives an overview of our approach.

\begin{figure}[t]
\centering
\includegraphics[scale=0.25]{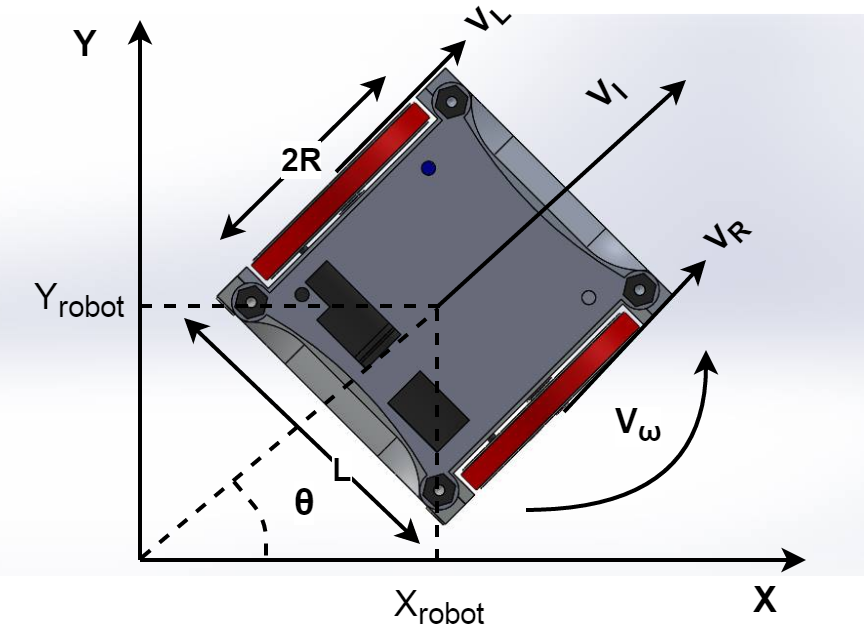}
\caption{Robot Kinematic Model. $(x,y,\theta,v_l,v_w)$ constitute robot state, which are governed by standard model for differential robots with $ x' = v_{l}cos(\theta), y' = v_{l}sin(\theta), \theta' = v_{\omega}$.}
\label{fig:kinematics}
\end{figure}

Another interesting property of Bayesian optimisation is the use of a surrogate function. Surrogate function forms the prior, and can be used to guide optimisation to regions containing minima points. We reuse this prior as obtained by Bayesian optimisation of trajectories, and store it in an offline reference database. This reduces computation time since many similar situations arise in robot soccer. We use starting and ending positions of the robots, their velocities, and obstacle positions as features for training k-Nearest Neighbour~\cite{references:cover1967nearest} model. The k-Nearest Neighbour technique identifies the closest prior from the reference database, which is reused to reduce optimisation time. We test our approach\footnote{https://github.com/abhinavagarwalla/motion-simulation} on simulator developed by us. We also experimented with robots designed and developed by Kharagpur RoboSoccer Students' Group\footnote{http://www.krssg.in} (KRSSG), and report significant improvements in the computation of trajectories as compared to the state of the art techniques. The kinematics of the manufactured robot is depicted in Fig. \ref{fig:kinematics}.

Our main contributions summarized:
\begin{itemize}
\item We introduce the use of Bayesian optimisation for determining control points leading to time-efficient trajectories that avoid obstacles, and kinematic constraints of the differential drive robots.
\item We reuse prior information of surrogate model in Bayesian optimisation from previously optimised trajectories to reduce trajectory optimisation time.
\item We identify most suitable prior to be reused for a particular situation using k-Nearest Neighbours algorithm~\cite{references:cover1967nearest}.
\item We extensively test our approach on a simulator as well as on physical soccer playing robots.
\end{itemize}

\section{Related Work}
Traditional approaches such as PolarBased~\cite{references:de2001control}, MergeSCurve~\cite{references:nguyen2008algorithms} and Dynamic Window~\cite{references:brock1999high}~\cite{references:seder2007dynamic} have several pitfalls, most important to us being: (1) sub-optimal traversal due to being reactive in nature, (2) no guarantees and estimates of travel time and distance leading to erroneous high level behaviours, (3) no incorporation of kinodynamic constraints of the bot (except Dynamic Window) and (4) frequent slipping of wheels at high speed. Trajectory based methods generally use B\'{e}zier curves~\cite{references:jolly2009bezier}~\cite{references:sahraei2007real}, B-splines~\cite{references:shiller1991dynamic}  or Quintic B\'{e}zier Splines~\cite{references:lau2009kinodynamic} for trajectory generation along with building velocity profiles that incorporate kinodynamic constraints. Building on work by Shiller and Gwo~\cite{references:shiller1991dynamic}, and Lau \textit{et al.}~\cite{references:lau2009kinodynamic}, we use cubic B\'{e}zier splines for trajectory generation. We optimise it globally using Bayesian Optimisation~\cite{references:martinez2014bayesopt}.

Potential Field~\cite{khatib1986real} and Visibility Graphs~\cite{kunigahalli1994visibility} are widely used algorithms for obstacle avoidance. While the former minimizes a potential function with  obstacles as repulsive and final goal as attractive poles, the optimisation procedure does not guarantee a globally optimal path. The latter generates paths that are very close to the obstacles frequently leading to collisions. Particle Swarm Optimisation~\cite{saska2006robot,wang2006obstacle} optimises Ferguson cubic spline trajectories, but are slow in real time because of large search space. On the other hand, our approach generates obstacle free optimised trajectories in real time.

Bayesian optimisation has been employed for planning, sensing and exploration tasks in robotics. In ~\cite{references:martinez2007active}, ~\cite{references:martinez2009bayesian} and ~\cite{references:souza2014bayesian}, authors propose an active learning algorithm for online robot path planning and reducing uncertainty in its state and environment under a time constraint. They essentially model the trade-off between exploration (uncertainty in environment) and exploitation (optimising planned trajectories) using a Gaussian process. In ~\cite{references:lizotte2007automatic} and ~\cite{references:calandra2016bayesian}, authors address the problem of gait optimisation for speed and smoothness on a quadruped robot. In a more machine learning context, ~\cite{references:bardenet2013collaborative} proposes a generic method for efficient tuning of new problems using previously optimised tasks. They use a novel Bayesian optimisation technique through combining ranking and optimisation. ~\cite{references:hoos2014efficient} analyses the effect of hyper-parameters on performance through a functional ANOVA framework imposed on RandomForest~\cite{liaw2002classification} predictions. Similar technique is discussed in ~\cite{references:brendel2011instance}, where authors map features and optimal parameters using an Artificial Neural Network, and then optimise using Covariance Matrix Adaptation Evolution Strategy \cite{references:hansen2003reducing}. We extend these approaches in a kinodynamic planning framework using splines through reuse of prior information from Bayesian optimisation.


Rosman \textit{et al.}~\cite{references:rosman2016bayesian} formalise Bayesian Policy Reuse (BPR) as solving a task within a limited number of trials by a decision making entity equipped with a library of policies. BPR introduces the problem where multiple policies are learnt for solving a single task in a reinforcement learning context. In contrast, we only store a single policy for different tasks of planning in a robot soccer domain tackling planning. We restrict the more general concept of BPR to policy reuse of Bayesian optimisation to suit our use case.

This paper is organised as follows: Section $3$ details trajectory optimisation process using Bayesian optimisation and database formation for prior reuse. Section $4$ introduces trajectory generation using splines and velocity profiling approach for kinematic constraints. Section $5$ describes the experimental setup of the simulator and kinematics of the manufactured robots. Section $6$ compares results of our approach with various other approaches, is followed by concluding Section $7$.

\section{TRAJECTORY OPTIMISATION}

Optimised trajectories improve winning chances by increasing ball possession and decreasing trajectory traversal time. However, optimising trajectories in a soccer match severely limits the reaction time of the robots due to computational overheads. We split trajectory optimisation into offline and online computation. First, trajectories are optimised offline and stored in a database. For online optimisation, the database is queried for similar situations to optimise it further. The proposed method is able to optimise trajectories under strict real time constraints.

We employ Bayesian optimisation technique for both offline and online optimisation of positions of control point. Setting control points at different locations leads to different trajectories, as described in Section 4. Traversal time for the generated trajectories is taken as the objective to be minimised. Apart from time, avoiding wheel slipping and high curvature paths is also important. These are handled implicitly in the trajectory generation process as described in Section 4. Formally, the objective $F$ is set to minimise traversal time depending on the position of control points of the generated trajectory. Control points are positions through which a trajectory must pass, and can be interpreted as parameters for trajectory generation. These are denoted using $CP_{i,x}$ and $CP_{i,y}$ where $i$, $x$ and $y$ refer to $i^{th}$ control point and coordinate axis. $j$ denotes the total number of control points. The set of control points is denoted by $P$, where $P$ = $\{CP_{1,x},CP_{1,y} \dots CP_{j,x},CP_{j,y}\}$. $P$ forms the parameter set for optimisation.
\begin{equation}
\begin{split}
    F(P) = F(\{CP_{1,x},CP_{1,y} \dots CP_{j,x},CP_{j,y}\})= \\\text{Traversal time of generated trajectory}
\end{split}
\end{equation}

\subsection{Bayesian Optimisation}
Since $F$ is a computationally expensive non-linear function for which closed form computation or gradient calculation is not possible, Bayesian optimisation~\cite{references:martinez2014bayesopt} is employed to minimise $F$.
\begin{equation}
P^{*} = \argmin_{P} F(P)
\end{equation}
Bayesian optimisation is a surrogate based global optimisation method that tries to optimise with very few function evaluations. As opposed to other optimisation methods, it remembers the history of evaluations i.e. acquired samples on which objective has already been evaluated. Using this history and an acquisition function, it decides the next sampling point $P$ to evaluate $F$ on. It can efficiently handle exploration-exploitation dilemma by not only modelling uncertainty in surrogate model, but also predicting value of objective $F$ at the next sampling point. The next sampling point is selected such that it minimises the uncertainty in the functional space of the objective.

Bayesian optimisation effectively shifts costly evaluations of objective function to cheap evaluations of the surrogate model $G$. Gaussian Process~\cite{references:rasmussen2006gaussian} is utilised as the surrogate model. The advantage of using Gaussian Process as surrogate is that a closed form solution for acquisition function is obtainable since Gaussian distribution is a self-conjugate distribution. Gaussian process models the function $f: P \rightarrow F(P)$, through a mean function and a covariance function $k(A, B)$, where $A$ and $B$ denote two parameter sets. We fix mean as a constant $m$, and use Automatic Relevance Determination (ARD) Mat\'{e}rn 5/2 kernel~\cite{references:snoek2012practical} as the covariance function. ARD Mat\'{e}rn 5/2 kernel, denoted by (\ref{eq:kernel}) avoids overly smooth functions, as generated by primitive kernels. $\sigma_s$ denotes covariance amplitude, and is generally kept as 1. In (\ref{eq:kernel2}), $A_i$, $B_i$ and $l_i$ denote $i$th parameter for set $A$ and $B$, and characteristic length-scale factor respectively. $n$ denotes the number of parameters, and $d$ captures the distance between two parameter sets.
\begin{equation}
\label{eq:kernel}
k(A, B) = \sigma_s^2(1+\sqrt{5d}+\frac{5}{3}d)\exp(-\sqrt{5d})
\end{equation}
\begin{equation}
\label{eq:kernel2}
d = \sum_{i=1}^{n}\frac{(A_i - B_i)^2}{l_{i}^2}
\end{equation}

Given a history of evaluated points, the next evaluation point is selected as the one which maximises the acquisition function. We use Expected Improvement~\cite{references:snoek2012practical} as acquisition function $C$, given by
\begin{equation}
C_{EI}(P) = \sigma(P)(u \Phi(P) + \phi(u))
\end{equation}
\begin{equation}
u = \frac{F(P_{best}) - \mu(P)}{\sigma(P)} 
\end{equation}
where $\mu$ and $\sigma$ denote the mean and variance of objective function as estimated by Gaussian Process model at $P$. $\mu$ and $\sigma$ map control point set $P$ to a real number signifying the estimate of the value and variance of the objective function respectively. $\Phi$ and $\phi$ denote the cumulative standard normal function and standard normal function in one dimension respectively. Bayesian optimisation has its own set of hyperparameters, namely length-scaling factors $l_i$, covariance amplitude $\sigma_s$, observation noise, and kernel parameters $u$ and $\sigma$. These are optimised by maximising the marginal likelihood of Gaussian process, as discussed in ~\cite{references:bergstra2011algorithms}.

\begin{table}[h]
\caption{Bayesian Optimisation parameters}
\label{params}
\begin{center}
\begin{tabular}{|c|c|}
\hline
Parameter & Value \\
\hline
Number of dimensions: n & $\mid P\mid$ \\
Range & [-1000, 1000] \\
Objective: $F(P)$ & Equation (1) \\
Acquisition: $C(P)$ & Expected Improvement\\
Kernel: $K(A, B)$ & ARD Matern 5/2 kernel \\
Prior: $P(F)$ & $\mathcal{GP}(0,{K(\theta=\{\mathit{l_i\dots,l_n},\sigma_s\})}+\sigma_{n}^2I)$ \\
\hline
\end{tabular}
\end{center}
\end{table}

An initial set $X$ of points are sampled using Latin hypercube sampling~\cite{references:stein1987large} which generates random numbers from a stratified input distribution. The surrogate model $G$ is incrementally updated to encode the observation history, from which the point maximising the acquisition function is selected as next evaluation point. The process is repeated until a specified number of evaluations or limited computation time. We experiment with various kernel functions, namely Mat\'{e}rn 5/2 kernel, Mat\'{e}rn 3/2 kernel and Squared Exponential kernel~\cite{references:snoek2012practical}. Various acquisition functions such as Expected Improvement~\cite{references:schonlau1998global}, A-optimality criteria and Lower Confidence Bound~\cite{references:cox1992statistical}. The chosen design parameters for Bayesian optimisation is summarised in Table~\ref{params}.

\newcommand\mycommfont[1]{\footnotesize\ttfamily\textcolor{black}{#1}}
\SetCommentSty{mycommfont}

\begin{algorithm}
\label{offline}
\SetAlgoLined
\KwData{Start position $SP(x, y)$, Ending position $EP(x, y)$, Starting velocity $SV(x, y)$, Ending velocity $EV(x, y)$, Obstacle position $OP(x, y)$, Number of control points $j$}
\KwResult{$D$ is the database}
 $x = (SP, EP, SV, EV, OP)$\;
 $P = {\{CP_{1,x},CP_{1,y} \dots CP_{j,x},CP_{j,y}\}}$\;
 $K(P_1,P_2)$: Kernel Function\;
 $C(P)$: Acquisition Function\;
 $N$: Number of iterations\;
 $\{X, y\} = \phi$\;


 \While{$iter$ $<$ $N$}
 {
 	$\theta$ = Bayesian optimisation hyperparameters\;
    Posterior Model Update: $Pr(F|D) \propto \int Pr(D|F,\theta)Pr(F)Pr(\theta)d\theta$\;
    Select $P_i$ $=$ $\underset{P}{\operatorname{argmax}}$ $C(P|Pr(F|D))$\;
    $\{X,y\} = \{X,y\} \cup \{P_i,F(P_i)\}$\;
 }
 Best $P^*=$ $\underset{P}{\operatorname{argmax}}$ $y$\;
 $B^*$ = getTrajectory($P^*$, SP, EP, SV, EV, OP) \tcp*{refer Section $4$}
Update database D with input $x$, $P^{*}$ and GP Model as prior
 \caption{Trajectory optimisation and Database formation using Bayesian optimisation}
\end{algorithm}

\begin{algorithm}
\label{alg:online}
\SetAlgoLined
\KwData{Start position $SP(x, y)$, Ending position $EP(x, y)$, Starting velocity $SV(x, y)$, Ending velocity $EV(x, y)$, Obstacle position $OP(x, y)$, Database $D$}
\KwResult{$B^{*}$ is the optimal trajectory}
\caption{Run-time Bayesian optimisation}
$x = (SP, EP, SV, EV, OP)$\;
\For{i=1 to m} 
{
 	Compute distance $d(D_i, x)$\;
}
Compute set $I$ containing indices for the $k$ smallest distances\;
Query D to get $k$ priors\; 
Average $k$ priors\;
Run Algorithm 1, and get optimised trajectory $B^{*}$\;
Return $B^{*}$\;
\end{algorithm}

During a robot soccer game, the generated trajectories should be safe i.e. it should have a good distance with obstacles to avoid collisions. In our case, obstacle avoidance is simply achieved through adding a penalty to the objective function if the current trajectory intersects an opponent robot. This results in the optimisation step rejecting the trajectories that collide with the obstacles. The field boundaries are also taken as line obstacles, resulting in rejection of trajectories with any point outside the soccer field.


\subsection{Prior Database Creation}
In a robot soccer match, trajectories must be generated within a time window of a few hundred milliseconds. Despite being sample efficient, Bayesian optimisation is impractical for real time optimisation. Since we use Gaussian process as the surrogate model, both prior and posterior are Gaussian distributions. To reduce trajectory generation time, we reuse prior from offline optimised trajectories for online optimisation. Assuming that we have enough data points of optimised trajectories, prior is rich with areas of sample points that lead to minimisation of objective function. This results in a surrogate model which has lower uncertainty about the objective function surface. It efficiently trades off exploration and exploitation, leading to reduced function evaluations and trajectory generation time.

We form a database by optimising trajectories through Bayesian optimisation for various combinations of starting and ending positions of our robots and obstacles, summarised in Algorithm~1. We also vary planning scenario with the starting and ending velocity, number of obstacles and number of control points which greatly influence the generated trajectory. Trajectories generated for different planning scenarios are optimised offline using a simulator. For each trajectory, the database stores the optimal placement of control points as well as the prior parameters. Prior parameters are mainly constituted by surrogate model along with its mean function and kernel function. Other parameters such as variance in observation noise, and acquisition function are also utilised. Planning scenario comprising of starting and ending positions, velocities, number of obstacles, number of control points and obstacle positions are also stored. In this study, we simulate a total of $20,000$ random planning scenarios, and save them to the database. 

\subsection{Prior Reuse}
With numerous starting and ending positions of five enemy obstacles and our bots, the space of possible configurations is really huge. This makes storing all situations computationally prohibitive facilitating the need to reuse previously optimised trajectories. Also, offline optimisation is done on a simulator and not on physical robots. A lot of factors such as lighting conditions, wheel mount, playing surface, etc. can alter playing conditions. Thus, online optimisation is necessary to adapt from simulator environment to current playing conditions. 

After offline optimisation, we have a database which can be queried for identifying similar planning scenarios. A planning scenario is characterised through starting and ending positions, velocities, number of obstacles and obstacle positions. These characteristics are used as features for training a k-Nearest Neighbour model~\cite{references:cover1967nearest}. We use k-NN since it is a non-parametric technique that doesn't require learning of model parameters. Also, the model is not very sensitive to hyperparameters which eliminates the need to look for optimum hyperparameters.

The k-Nearest Neighbour (k-NN) technique is based on constructing a space partitioning tree on the training dataset. The constructed tree can be searched at testing time to identify data points similar to test query. It uses a distance function to identify the similarity between the test query point and the training dataset. Many distance functions can be utilised for measuring the similarity between two $n$-dimensional instances $A, B$. We utilise L1-distance $d$, as denoted by (\ref{eq:distance}), as the distance function in this study. $A_{i}$ and $B_{i}$ denotes the component of $A$ and $B$ along $i^{th}$ dimension.
\begin{equation}
\label{eq:distance}
d(A, B) = \sum_{i=1}^{n}\mid A_{i} - B_{i}\mid
\end{equation}
Our method is general in the sense that other sophisticated machine learning models that employ a notion of similarity can also be employed. The model identifies the most similar planning scenario and its corresponding prior using k-NN. The prior comprising of the mean function and kernel function of surrogate model is used to initialise Bayesian optimisation. It then optimises trajectory in very few steps until either convergence is achieved or computation time runs out. Algorithm~2 summarises prior reuse for real time Bayesian optimisation.


\section{TRAJECTORY GENERATION}
In this section, we explain how a trajectory is generated and traversal time is computed. Given the starting point, ending point and control points, trajectory are parametrized by splines. Velocity profiling algorithm ensures the kinematic constraints of the bot, and profiles velocity vector at each point on the trajectory. Trajectory traversal time is calculated by accumulating time and velocity attained between two successive points. A tracking algorithm ensures that the robot does not deviate from the generated path.

Cubic B\'{e}zier Splines~\cite{references:farin2014curves} are chosen to represent trajectories because of their desirable smoothness properties. Two dimensional paths are obtained by parametrization of path along $x$ and $y$ directions through the parameter $u$ along the curve.
Initial trajectory is generated by stitching together $n$ segments of Cubic B\'{e}zier Splines along ${n-1}$ control points. The equation for the Cubic B\'{e}zier Splines is given by (\ref{eq:1}), which follows the continuity and differentiability constraints at the knots in (\ref{eq:2}), (\ref{eq:3}) and the boundary conditions in (\ref{eq:4}). Initial robot velocity which is captured from the overhead camera is incorporated in (\ref{eq:5}) resulting in formation of end slope cubic splines. $B_i$ denotes $i^{th}$ segment of the spline with $P_{i0}$, $P_{i1}$, $P_{i2}$ and $P_{i3}$ as coefficients.

\begin{equation} \label{eq:1}
B_i (u) = 
\begin{bmatrix}(1 - u)^{3} \\3(1 - u)^{2}u \\3(1 - u)u^{2} \\u^{3} \end{bmatrix}^T
\begin{bmatrix}P_{i0} \\P_{i1} \\P_{i2} \\P_{i3}\end{bmatrix}
\end{equation}
\begin{equation} \label{eq:2}
B_i(1) = B_{i+1}(0)
\end{equation}
\begin{equation} \label{eq:3}
B_i^{'}(1) = B_{i+1}^{'}(0), B_i^{''}(1) = B_{i+1}^{''}(0)
\end{equation}
\begin{equation} \label{eq:4}
B_i(0) = P_{i0}, B_i(1) = P_{i3}
\end{equation}
\begin{equation} \label{eq:5}
B_i^{'}(0) = v_l(0), B_i^{'}(1) = v_l(1)
\end{equation} 

Quintic B\'{e}zier Splines~\cite{references:lau2009kinodynamic} which have the additional property of curvature continuity at end points, are compared with their cubic counterparts in terms of trajectory traversal time, tracking error and average velocity of the robot. Our method is inherently bidirectional by taking account of the direction of velocity at the time of trajectory generation.

\subsection{Arc Length Re-parametrisation}
Arc length re-parametrisation is required to compute the parameter $u$ in (7), from the arc length $S$, which is equivalent to finding point on trajectory at which the arc length equals $S$. Re-parametrisation is implemented through B\'{e}zier splines~\cite{references:wang2002arc} and inverting B\'{e}zier curves~\cite{references:walter1996approximate}, of which former is finally chosen because it provides better performance. Equidistant points are sampled from $u \in {[0,1]}$, and arc length on each of these points is computed. Then, a 1D Cubic B\'{e}zier Spline is fitted on these points, approximating $u$ as a function of $S$. Now, getting the corresponding $u$ for a given arc length, is a simple evaluation of the cubic spline at $u$ using equation (\ref{eq:1}).

\subsection{Velocity Profiling}
Velocity profile must be generated incorporating robot dynamics to minimise the error between observed and expected position at any moment of time. While keeping the errors to minimum, velocity at each point is maximised while respecting the kinodynamic constraints. On the trajectory $B(u)$, we fix a set of $N_c$ equidistant planning points $p_{i}$ along the arc length using arc length re-parametrisation. Planning points are just sampled from the generated trajectory. We have assumed that the acceleration between two planning points $p_{i}$ and $p_{i+1}$ is constant, which holds true for most trajectories in practice due to selection of sufficiently many points. Each translational velocity $v_{i}$ corresponds to a planning point $p_{i}$. Initial velocity $v_0$ is set to initial velocity of robot, and final velocity $v_{N_c}$ is set according to game-play. Forward consistency is established by accelerating with the maximum constrained acceleration from point $p_{i}$ to reach $p_{i+1}$ with maximum velocity. The process is repeated in the reverse direction. This ensures that the robot traverses between any two points obeying the constraints. The total time needed to complete the trajectory is also obtained, once the velocity $v_i$ at each planning point $p_i$ is determined.

Assuming that the acceleration is constant between the two planning points $p_i$ and $p_{i+1}$, we compute the time taken to reach any of the equidistant planning points using (\ref{eq:time1}), where $v_i$ and $v_{i+1}$ are determined from velocity profiling. Time taken to reach planning points $p_i$ and $p_{i+1}$ is denoted by $t_{i}$ and $t_{i+1}$ respectively. $\Delta_{s}$ denotes the distance between $p_i$ and $p_{i+1}$. The time at the end point $N_c$ is given by (\ref{eq:time2}).
\begin{equation}\label{eq:time1}
    t_{i+1} = t_{i} + \frac{2\Delta_{s}}{v_i + v_{i+1}}
\end{equation}
\begin{equation}\label{eq:time2}
    totalTime = t_{N_c}    
\end{equation}

\subsection{Constraints}
The kinodynamic constraints limit the maximum velocity $v_{i}$ that can be set at each planning point $p_{i}$. The maximum translational velocity $v_{max|\omega}$ for angular velocity $\omega$ at a point $p_{i}$ is given by (\ref{eq:6}), where $\omega_{max}$ and $\kappa_{i}$ denote maximum rotational velocity and curvature at that point, respectively.
\begin{equation} \label{eq:6}
    v_{i} \in [0, v_{max|\omega}], {v_{max|\omega}} = \frac{\omega_{max}}{|\kappa_{i}|} 
\end{equation}
The acceleration constraints at any planning point $p_i$ depends on the velocities of the adjacent points $p_{i-1}$ and $p_{i+1}$. The acceleration constraints are symmetric for the motion planning to be smooth and to avoid slipping of the robot. At any planning point $p_{i}$, $v_{i} \in [v_{min|a_{t}}, v_{max|a_{t}}]$ for acceleration $a_{t}$ can be attained using (\ref{eq:7}), and (\ref{eq:8}) where $a_{t_{max}}\Delta_{t}$ denotes maximum translation acceleration.
\begin{equation} \label{eq:7}
    v_{min|a_{t}} = v_{i - 1} - a_{t_{max}}\Delta_{t}
\end{equation}
\begin{equation} \label{eq:8}
    v_{max|a_{t}} = v_{i - 1} + a_{t_{max}}\Delta_{t}
\end{equation}
 The boundary conditions can now be obtained for translational velocities obeying  $a_{t_{max}}$, where distance between any two planning points $p_{i}$, $p_{i + 1}$ is $\Delta_{s}$. The reader can refer to~\cite{references:sprunk2008planning} for details.
\begin{equation} \label{eq:9}
    v_{min|a_{t}} = \begin{cases} 
                    \sqrt{v_{i - 1}^{2} - 2a_{t_{max}}\Delta_{s}} & v_{i - 1}^{2}\ge 2a_{t_{max}}\Delta_{s} \\
                    0 & else
  \end{cases}
\end{equation}
\begin{equation} \label{eq:10}
    v_{max|a_{t}} = \sqrt{v_{i - 1}^2 + 2a_{t_{max}}\Delta_{s}}
\end{equation}

\subsection{Tracking}
The tracking module ensures that the robot follows the trajectory by minimising the error between the current robot state and the expected state at that moment of time. The current tracker is very similar to~\cite{references:klanvcar2007tracking}. Characteristic frequency $\omega_{n}(t)$ of the system is given by (\ref{eq:11}) where $v_r$, $\omega_r$ are tangential and angular velocity, respectively. Controller gains $k_1$, $k_2$, $k_3$ are given by (\ref{eq:12}) where $\zeta$ represents damping coefficient, and $g$ is simply a parameter. Errors in position $(x,y)$, orientation $\theta$, and feed forward tangential and angular velocity are denoted by $e_1$, $e_2$, $e_3$, $u_{r1}$ and $u_{r2}$ respectively. The final tangential velocity $v$ and angular velocity $w$ sent to the bot are given by (\ref{eq:13}) and (\ref{eq:14}) respectively.

\begin{equation} \label{eq:11}
    \omega_n(t) = \sqrt{\omega_r(t)^2 + gv_r(t)^2}
\end{equation}
\begin{equation} \label{eq:12}
    k_1 = k_3 = 2\zeta\omega_n(t),  
    k_2 = g|v_{r}(t)|
\end{equation}
\begin{equation} \label{eq:13}
    v = u_{r1}cos(e_3) + k_1e_1
\end{equation}
\begin{equation} \label{eq:14}
    w = u_{r2} + sgn(u_{r1})k_2e_2 + k_3e_3
\end{equation}




\section{Experimental Setup}

\begin{figure}[t]
\centering
\includegraphics[scale=0.2]{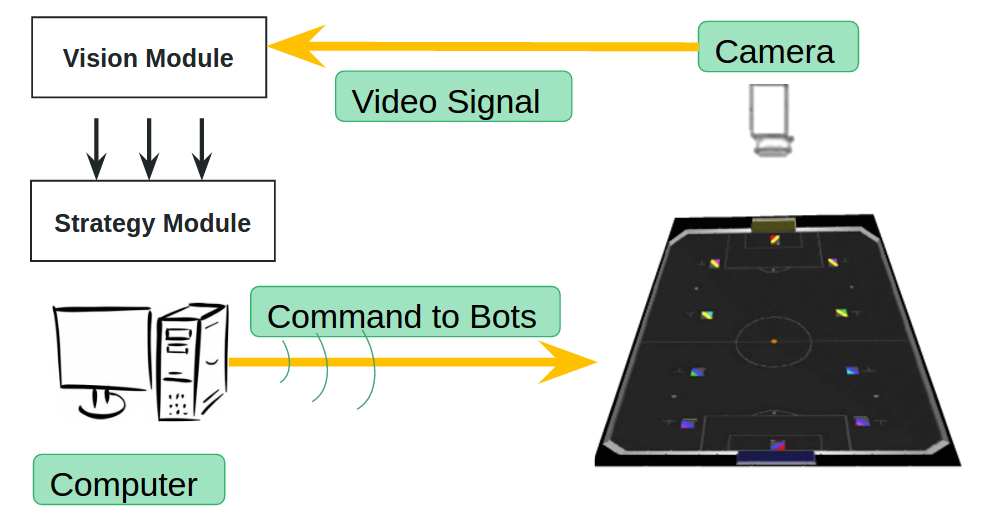}
\caption{System Setup and Overview: Raw camera feed is used by vision module to identify robot positions and orientations. Strategy module initiates trajectory generation process, after which velocity commands are sent to the robots.}
\label{fig:game}
\end{figure}

\subsection{Robot Dynamics}
Federation of International Robot-Soccer Association (FIRA)\footnote{https://www.fira.net/} organises Mirosot league every year, in which a team of five mobile robots compete against an opponent team to win a soccer match. A Mirosot robot is a differential drive robot with two wheels on the sides and supportive caster wheels on the front and back. The robot has zero turn radius, and is fitted a cube of side length 7.5cm. The system setup is described in Fig. \ref{fig:game}. Raw values obtained from camera are passed through Kalman~\cite{references:ribeiro2004kalman} filter at 60fps, thereby updating ball and robot state. An NRF module then transmits the velocities to the micro-controller fitted on each robot. There is an internal Fuzzy-PID~\cite{references:visioli2001tuning} loop running on the motors, which decides the velocity of motors.

\subsection{Physical Constraints}
Taking into account wheel dimensions, Fuzzy-PID control loop, circuit design and motors specifications, the maximum loaded velocity attainable by the robot is set to 200 cm/s. For calculating the radial acceleration, the robot is run on a circular path with a fixed radius as in ~\cite{references:lepetivc2003time}. The value of radial acceleration $a_{rad}$ is increased in steps till the slipping point. Using linear regression, radial acceleration is obtained as a function of radius of curvature $r$, given by (\ref{eq:15}). 
\begin{equation} \label{eq:15}
    a_{rad} = -5.92r + 700
\end{equation}

\subsection{Simulator}
A simulator is also developed for testing the locomotion algorithms. It is specially designed to enable simulation with kinematic and physical model of our robots. The dimensions of the simulated robots, and actuator limits are set as the same as that of actual robots. It discretises the transmission into steps of 16ms, accounts for packet delay and adds noise as encountered while working on real robots. The mobile robot does not receive the sent velocities instantaneously. There is a time lag between transmitting and receiving of packets, which we define as packet delay, and is calculated to be four packets i.e. 64ms. Due to the special care taken in designing the simulator, only minimal retuning of parameters is required and the same code works on both the simulator and the real robots. This has also proved useful in pre-tuning parameters in simulator, and then tuning them on real robots.


\begin{figure*}[ht]
  \begin{subfigure}[b]{0.25\linewidth}
    \includegraphics[width=0.95\linewidth,height=1.2in]{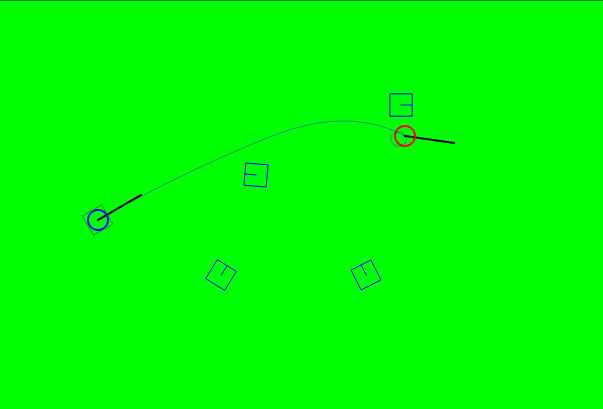} 
  \end{subfigure}
  \begin{subfigure}[b]{0.25\linewidth}
    \includegraphics[width=0.95\linewidth,height=1.2in]{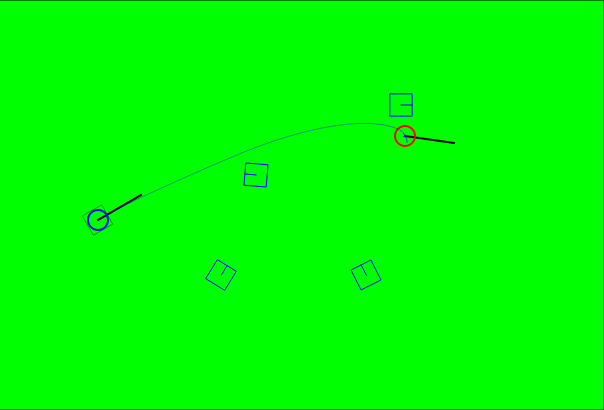} 
  \end{subfigure}
  \begin{subfigure}[b]{0.25\linewidth}
    \includegraphics[width=0.95\linewidth,height=1.2in]{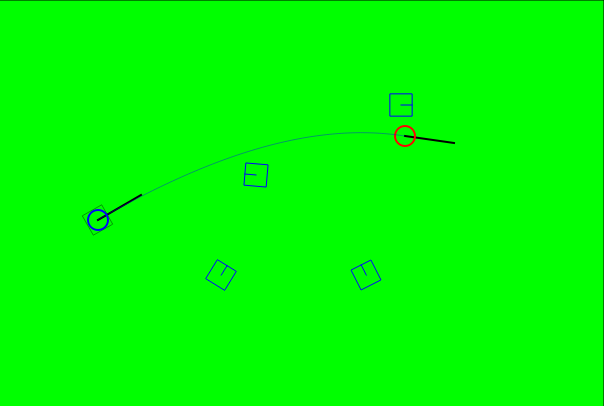} 
  \end{subfigure}
  \begin{subfigure}[b]{0.25\linewidth}
    \includegraphics[width=0.95\linewidth,height=1.2in]{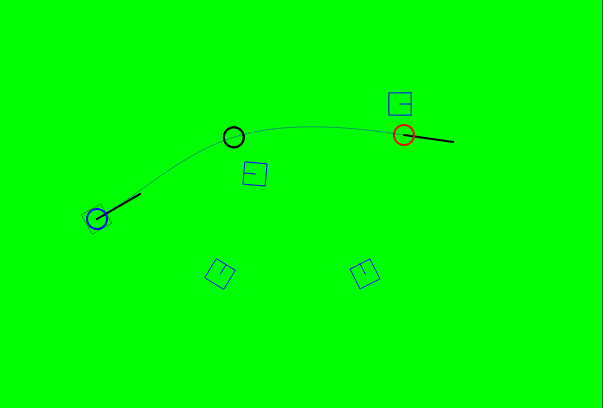} 
  \end{subfigure}
  
  \vspace{0.05in}
  \begin{subfigure}[b]{0.25\linewidth}
    \includegraphics[width=0.95\linewidth,height=1.2in]{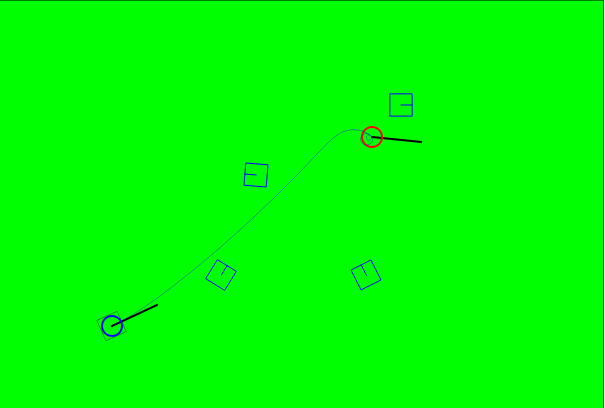} 
  \end{subfigure}
  \begin{subfigure}[b]{0.25\linewidth}
    \includegraphics[width=0.95\linewidth,height=1.2in]{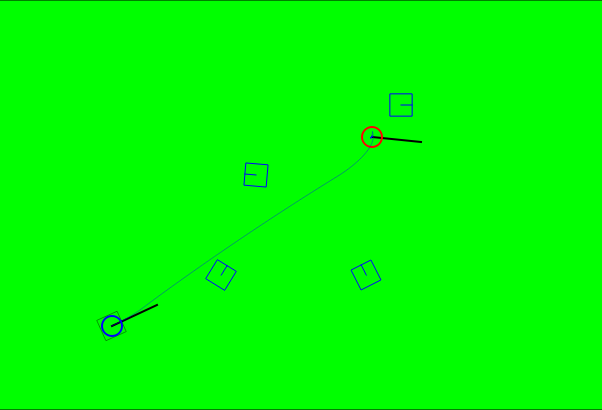} 
  \end{subfigure}
  \begin{subfigure}[b]{0.25\linewidth}
    \includegraphics[width=0.95\linewidth,height=1.2in]{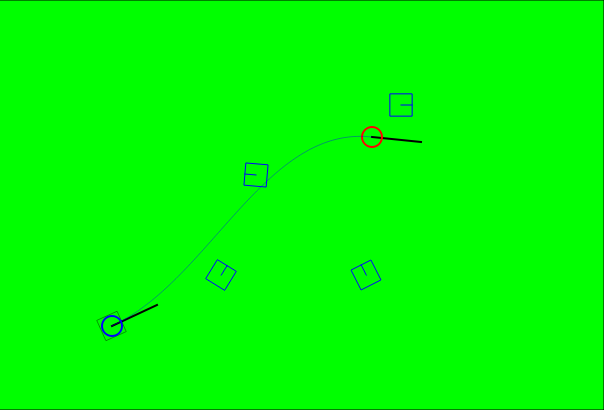} 
  \end{subfigure}
  \begin{subfigure}[b]{0.25\linewidth}
    \includegraphics[width=0.95\linewidth,height=1.2in]{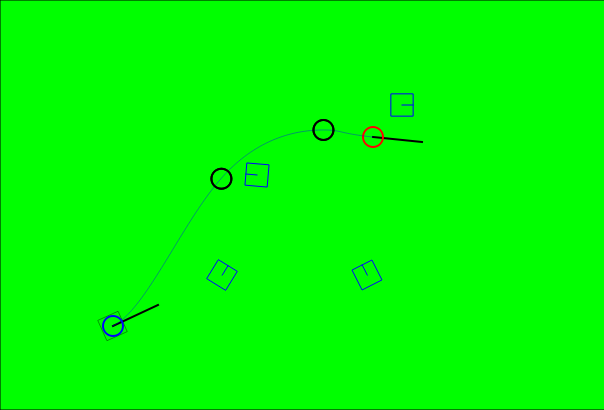} 
  \end{subfigure}
  
  \vspace{0.05in}
  \begin{subfigure}[b]{0.25\linewidth}
    \includegraphics[width=0.95\linewidth,height=1.2in]{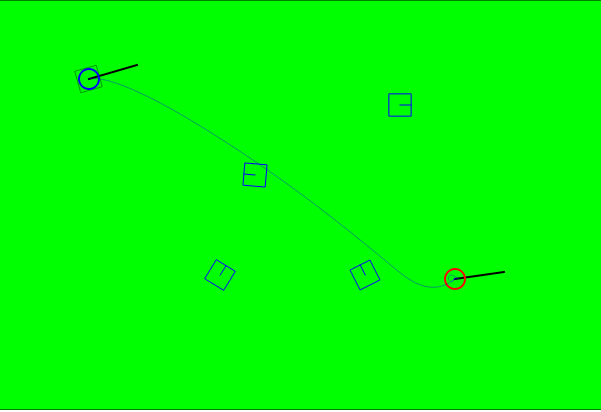} 
  \end{subfigure}
  \begin{subfigure}[b]{0.25\linewidth}
    \includegraphics[width=0.95\linewidth,height=1.2in]{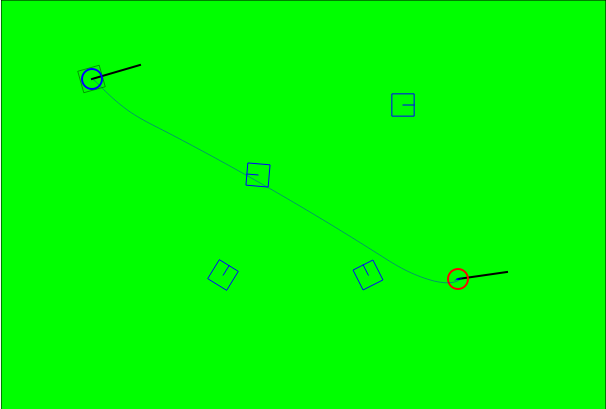} 
  \end{subfigure}
  \begin{subfigure}[b]{0.25\linewidth}
    \includegraphics[width=0.95\linewidth,height=1.2in]{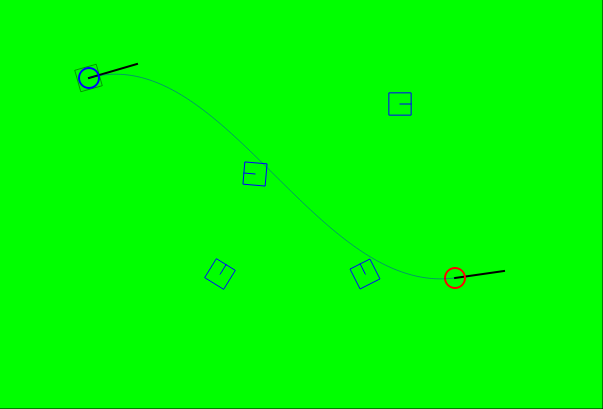} 
  \end{subfigure}
  \begin{subfigure}[b]{0.25\linewidth}
    \includegraphics[width=0.95\linewidth,height=1.2in]{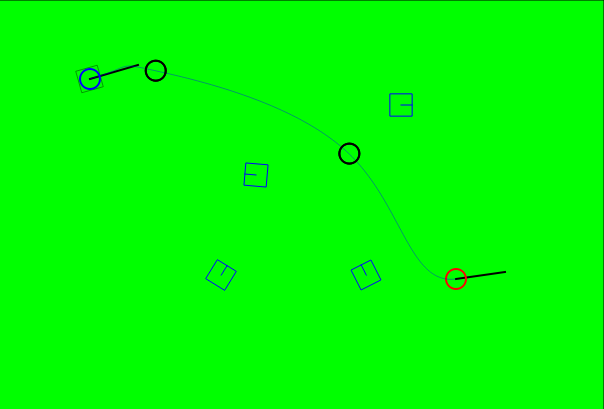} 
  \end{subfigure}
  
  \vspace{0.05in}
  \begin{subfigure}[b]{0.25\linewidth}
    \includegraphics[width=0.95\linewidth,height=1.2in]{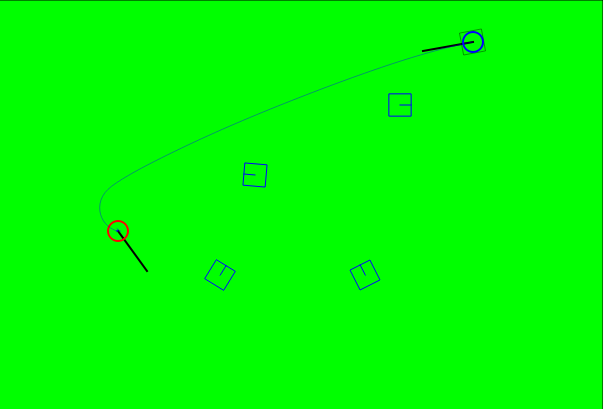} 
  \end{subfigure}
  \begin{subfigure}[b]{0.25\linewidth}
    \includegraphics[width=0.95\linewidth,height=1.2in]{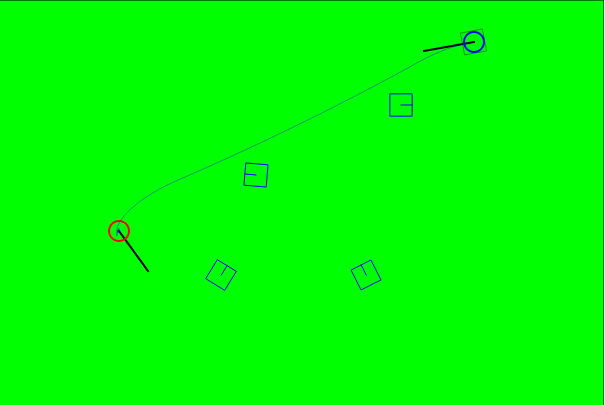} 
  \end{subfigure}
  \begin{subfigure}[b]{0.25\linewidth}
    \includegraphics[width=0.95\linewidth,height=1.2in]{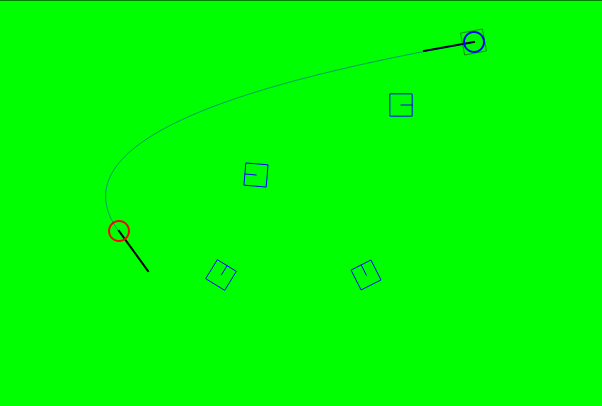} 
  \end{subfigure}
  \begin{subfigure}[b]{0.25\linewidth}
    \includegraphics[width=0.95\linewidth,height=1.2in]{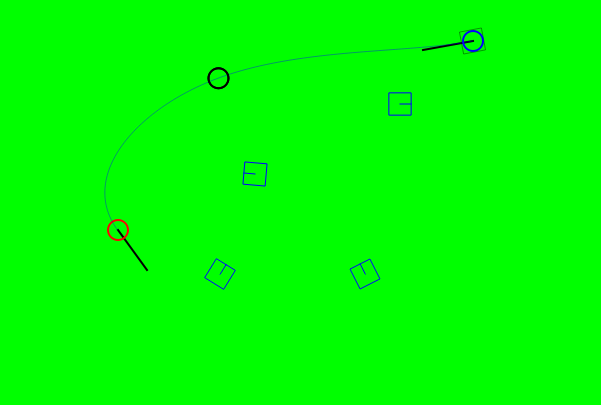} 
  \end{subfigure}
  
  \vspace{0.05in}
  \begin{subfigure}[b]{0.25\linewidth}
    \includegraphics[width=0.95\linewidth,height=1.2in]{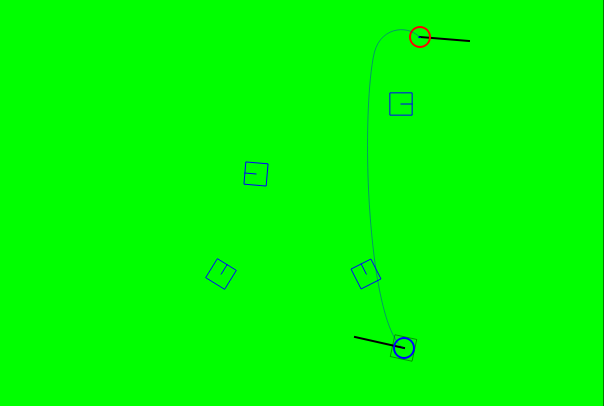} 
    \caption{S-curve}
  \end{subfigure}
  \begin{subfigure}[b]{0.25\linewidth}
    \includegraphics[width=0.95\linewidth,height=1.2in]{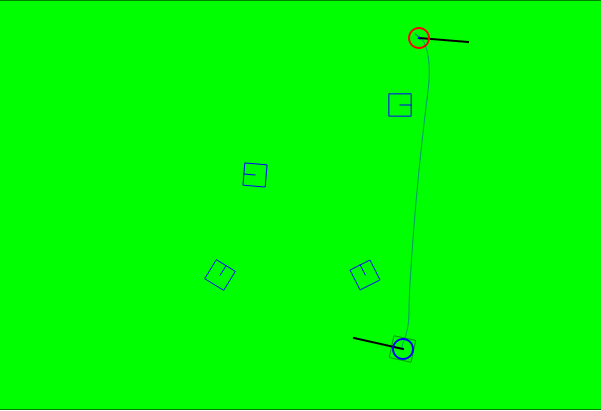} 
    \caption{Dynamic Window}
  \end{subfigure}
  \begin{subfigure}[b]{0.25\linewidth}
    \includegraphics[width=0.95\linewidth,height=1.2in]{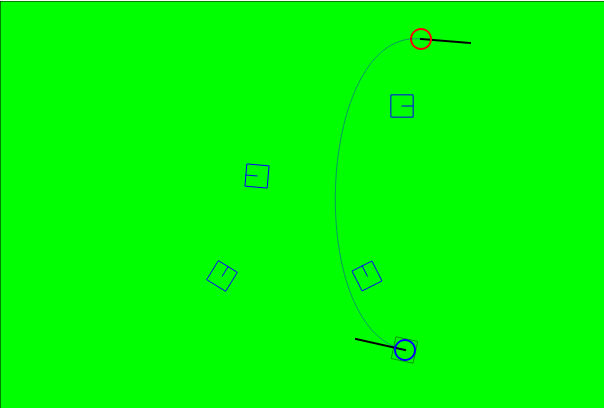} 
    \caption{PSO}
  \end{subfigure}
  \begin{subfigure}[b]{0.25\linewidth}
    \includegraphics[width=0.95\linewidth,height=1.2in]{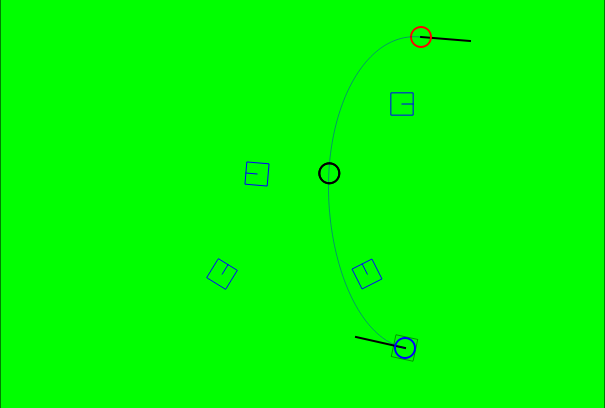} 
    \caption{Our method}
  \end{subfigure}
  \caption{Trajectory generated by a) S-curve, b) Dynamic Window, c) PSO and d) Our Method, for a set of start and end positions on our developed simulator. The blue circle and red circle represent the start and end positions respectively. Black line from the centre of the circles displays the orientation of the robot at those points, and obstacles are shown as blue squares. Trajectory is denoted by a blue curve between start and end points. In case of Our Method, the position of generated control points are shown by black circles.}
  \label{fig:paths} 
\end{figure*}

\begin{table}[t]
\caption{A comparison of Bayesian Optimisation with five other algorithms in terms of average time(in sec), tracking error(calculated using (\ref{eq:16})) and average velocity(in cm/s) by running them on our robots. The best scores are bold-faced.} 
\label{time_results}
\begin{center}
\begin{tabular}{|c|c|c|c|}
\hline
\textbf{Planner} & \textbf{Time(s)} & \textbf{te} & \textbf{Avg. Velocity(cm/s)}\\
\hline
\hline
\hline
S-curve~\cite{references:nguyen2008algorithms} & 2.675 & 6.634 & 83.338\\
\hline
Polar-Bidirectional~\cite{references:de2001control} & 4.190 & 4.836 & 42.536\\
\hline
Dynamic Window~\cite{references:seder2007dynamic} & 2.800 & 7.067 & 85.049\\
\hline
PSO~\cite{saska2006robot} & 2.331 & 4.279 & 94.614\\
\hline
Quintic Spline~\cite{references:lau2009kinodynamic} & 2.472 & 4.631 & 90.470\\
\hline
Proposed approach & \textbf{2.231} & \textbf{4.024} & \textbf{96.903}\\
\hline
\end{tabular}
\end{center}
\end{table}

\section{Results and Discussion}
We analyse the performance of our approach on multiple experiments, and compare its performance with other competitive motion planners. In the first set of experiments, we utilise Cubic B\'{e}zier Splines~\cite{references:farin2014curves} for trajectory generation, and Bayesian optimisation with prior reuse for trajectory optimisation. Fig. \ref{fig:paths} shows the generated trajectory for various planners for different starting and ending points. The field setup is similar to a robot soccer match with over-crowding of obstacles. S-Curve~\cite{references:nguyen2007planning} algorithm generates smooth path but the trajectory has very low radius of curvature at some points. Due to kinematic constraints of the robot, the robots slow down at these points leading to increase in total traversal time. The path generated by Dynamic Window~\cite{references:seder2007dynamic} has got very sharp turns at the end points which cause slipping. Since the generated trajectory is very close to the obstacles, the robot collides with the obstacles while following the path. The path generated by PSO~\cite{saska2006robot} looks promising but it is computationally expensive with few high curvature points. There are some cases in Fig. \ref{fig:paths} where the trajectory is very close to the obstacles. In our experiments, the population of PSO contains 15 particles, and takes around 100 iterations to converge. Trajectory generated by splines after Bayesian Optimisation has smooth turns, and maintains good distance with obstacles at all points. Thus, our method leads to trajectories that only avoid obstacles, but also reduce trajectory traversal time.

We plan a fixed set of manoeuvres for the robot comprising of a set of 10 starting and ending positions. The task of the robot is to reach from starting point to end point in minimum time possible while avoiding slipping. The robot must also result in low tracking errors for high-level attacking and ball interception abilities. Tracking error $te$ is taken as the log of mean squared error given by (\ref{eq:16}), where $x_{i,p}$,  $y_{i,p}$, $x_{i,a}$, $y_{i,a}$ denote the planned and actual positions of the robot for $x$ and $y$ coordinate respectively. $n$ denotes the total length of the trajectory. The tracking error $te$ might become large either due to slipping of wheels, or due to collision with another robot.

\begin{equation} \label{eq:16}
    te = \log{\frac{\sum_0^1\sqrt{(x_{i,p}-x_{i,a})^2 + (y_{i,p}-y_{i,a})^2}}{n}}
\end{equation}

\begin{figure}[t]
\centering
\includegraphics[scale=0.3]{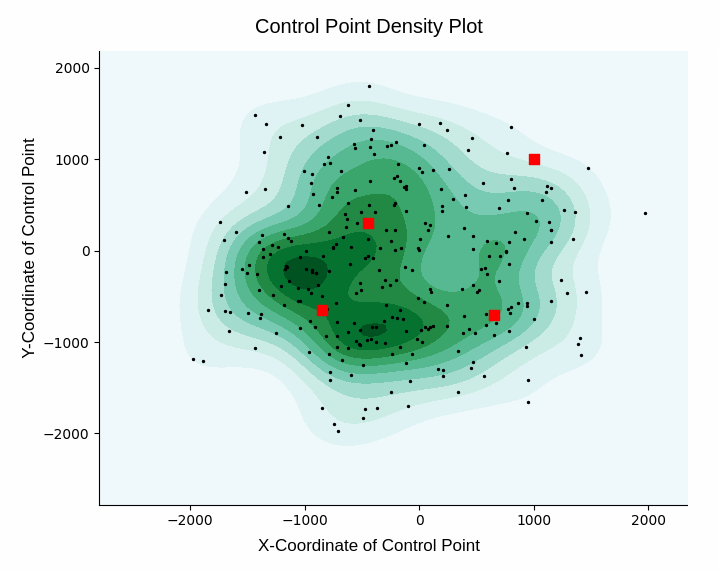}
\caption{Obtained control points as overlayed on a kernel density estimation plot for $300$ randomly initialised starting and ending points. Obstacles are depicted using red squares, and control points using black dots. In density plot, darker regions depict denser regions.}
\label{fig:bayes_density_cp}
\end{figure}

Table \ref{time_results} summarises the performances on the above task. We evaluate planners on average time taken to reach the destination, tracking error $te$ and average velocity of the robot while following the path. Our method performs better than other approaches on all the metrics. Online methods like Dynamic Window~\cite{references:seder2007dynamic} and S-curve~\cite{references:nguyen2007planning} show promising results in average time taken and average velocity, but have high tracking errors $te$ due to wheel slipping. The tracking error $te$ of Polar-Bidirectional~\cite{references:de2001control} planner is small compared to other online methods, but the velocity of the robot at each point is less leading to high traversal time. The Quintic Spline~\cite{references:lau2009kinodynamic} method results in sharp turns in some cases, which increase the average traversal time of the robot. Our method generates trajectories with high radius of curvature at all points, which helps the robot to move with higher speeds, and reduces the risk of slipping. Thus, the average speed is quite higher leading to less traversal time and tracking error $te$, when compared with other methods. 

It was also observed that on planning trajectories using S-curve, Polar-Bidirectional and Dynamic Window, the robot first reaches the end point at a high velocity, over shoots and keep oscillating around the end point. This property can also be seen by zooming into generated trajectories in Fig. \ref{fig:paths}. Also, these approaches didn't consider the orientation and velocity of the bot at end point, and reached the end point with an arbitrary velocity and orientation. This led to failure in correctly intercepting the ball, and poor ball handling performance.

\begin{figure}[t]
\centering
\includegraphics[scale=0.4]{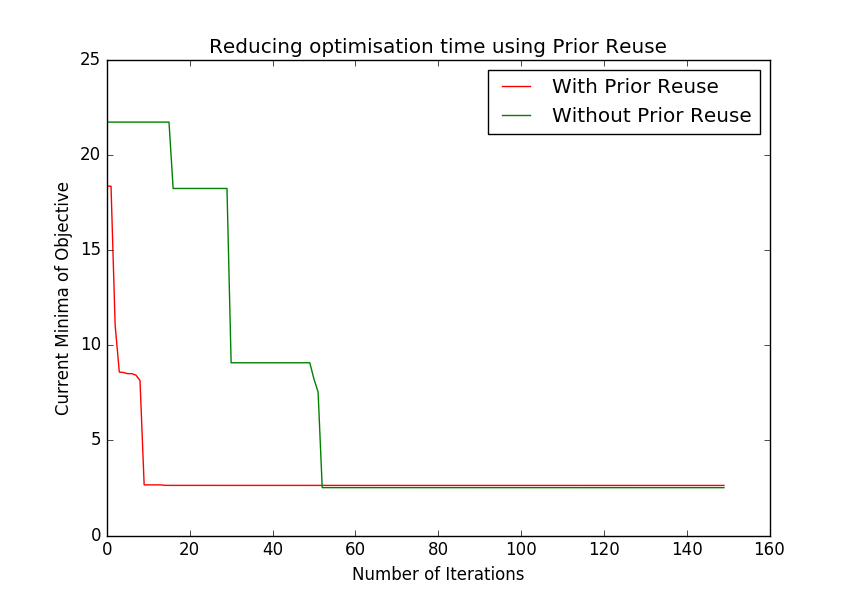}
\caption{Performance comparison of Bayesian Optimisation with (in red) and without (in green) reuse of prior information from database.}
\label{fig:online_1}
\end{figure}

We evaluate the computation time for online Bayesian optimisation for both cases, one involving prior reuse and other without prior reuse. Fig. \ref{fig:online_1} shows the number of iterations on the x-axis, and the current best minima of objective function on y-axis. For prior reuse, we utilise the closest prior as obtained from k-Nearest Neighbour to initialise the optimisation. It can be seen that prior reuse greatly reduces the number of iterations required to minimize the objective function. The optimisation process converges in around 15 iterations. Without prior use, it takes around 60 iterations to converge to the minima. Although the minima achieved without reusing prior is slightly better, but is insignificant considering that trajectory optimisation time is limited. This shows that our method is able to generate near optimal trajectories in a limited time setting.

Figure~\ref{fig:bayes_density_cp} shows the obtained control points for $300$ different combinations of starting and ending point with fixed obstacle positions. We see that the obtained control points avoid obstacles for all combinations, as no black point lies on the red square. The control points are concentrated in certain safe regions that are denoted by darker green regions. The generated trajectories not only avoid obstacles by traversing through the region in-between obstacles, but also reduce traversal time.

\begin{figure}[t]
\centering
\includegraphics[scale=0.4]{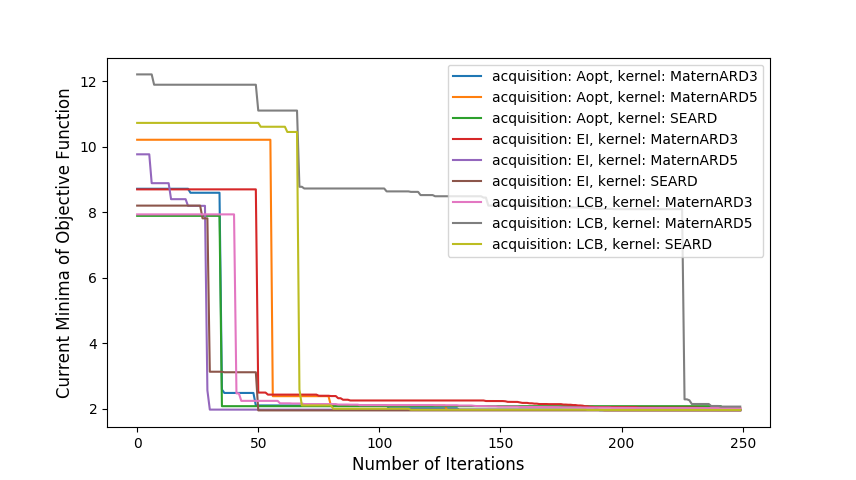}
\caption{Performance comparison in terms of number of iterations required to converge to minima for different combinations of kernel and acquisition function}
\label{fig:bayes_kernel_criterion}
\end{figure}

We also experiment with different hyperparameter settings for Bayesian optimisation. We evaluate the performance of different acquisition functions, namely Expected Improvement~\cite{references:schonlau1998global} (EI), Lower Confidence Bound~\cite{references:cox1992statistical} (LCB) and A-optimality criteria (Aopt). We also try different kernel function, namely Mat\'{e}rn 5/2 (MaternARD5), Mat\'{e}rn 3/2 (MaternARD5) and Square Exponential (SEARD) ~\cite{references:rasmussen2006gaussian}. We use Automatic Relevance Determination (ARD) for all these kernels. The trajectory travel times for all different combinations of kernel function and acquisition function are analysed for $20$ different starting and ending points. As shown in Table~\ref{bayesopt_kernel_criterion_results}, we see that different combinations lead to very similar results. The best results are obtained using either EI or LCB as acquisition function, along with MaternARD5 or MaternARD3 as kernel function. However, the number of iterations required to reach minima is different for these combinations, as shown in Fig.~\ref{fig:bayes_kernel_criterion}. We see that using MaternARD5 as kernel function, and EI as acquisition function takes the minimum number of iterations.

\begin{figure}[t]
\centering
\includegraphics[scale=0.3]{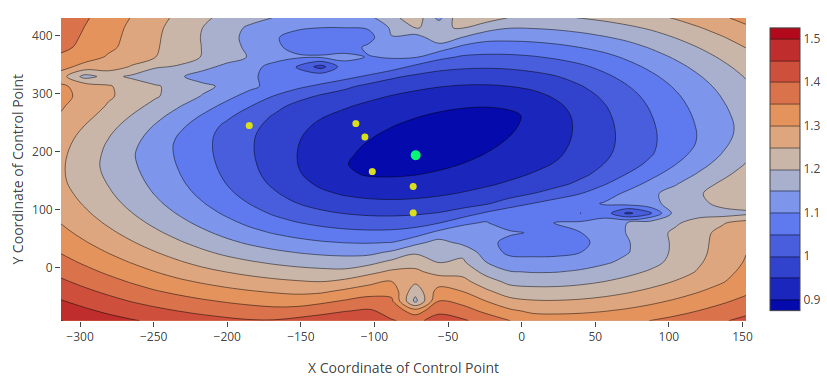}
\caption{Contour plot for trajectory traversal time (in seconds) generated using control points optimised using prior reuse. The bluer the region, lower the traversal time. Green point denotes control point obtained after Bayesian optimisation. The similar planning scenarios identified k-NN are shown in yellow.}
\label{fig:contour_knn}
\end{figure}

In the next experiment, we fix the starting and ending positions of our robot and the obstacles. Setting $k=6$, we query the database using k-NN, and initialise the prior for Bayesian optimisation. The prior is initialised by averaging the prior parameters across all the neighbours. We evaluate the performance of control points obtained after optimisation by measuring the trajectory traversal time for a dense grid surrounding the region. Figure~\ref{fig:contour_knn} shows the obtained contour plot, where bluer region denote regions minimising trajectory traversal time. We validate the use of k-NN for prior reuse as the control point obtained after optimisation minimises the objective.  
\begin{table}[t]
\caption{Trajectory traversal time (in seconds) for different combinations of kernel and acquisition function in Bayesian optimisation}
\label{bayesopt_kernel_criterion_results}
\begin{center}
\begin{tabular}{|c||c|c|c|}
\hline
\diagbox[width=7em]{\\ \textbf{Acquisition}}{\textbf{Kernel}} & \textbf{MaternARD3} & \textbf{MaternARD5} & \textbf{SEARD} \\
\hline
\hline
 \textbf{EI} & 1.437 & \textbf{1.436} & 1.452 \\
\hline
 \textbf{LCB} & 1.436 & \textbf{1.436} & 1.458 \\
\hline
 \textbf{Aopt} & 1.449 & 1.472 & 1.526 \\
\hline
\end{tabular}
\end{center}
\end{table}

\section{Conclusion}
Performing online optimisation under real time constraints is the key limitation of common optimisation methods, especially in complex robotics tasks. We combine learning and planning to address this problem in the context of trajectory optimisation in robot soccer. Our work utilises a database of previously optimised trajectories to effectively reduce optimisation time, and lead to time-efficient trajectories. Specifically, we store resultant prior by running Bayesian optimisation offline for different input configurations. At test time, the closest planning scenario is queried using k-Nearest Neighbours approach for reuse of prior. Bayesian optimisation is initialised with this prior, and run for few function evaluations. We test our method on multiple planning scenarios, where it outperforms traditional optimisation techniques.

We suggest that current optimisation technique are inadequate for real time optimisation for complicated tasks such as planning. If we are to make significant advances in this field, similar efforts targeting both theory and experimentation for combining learning and planning must be explored. Using more sophisticated machine learning algorithms that can better capture the relation between planning scenario and control point locations is left to future work. Querying multiple prior from the database, and combining them appropriately for a rich prior is also an interesting future direction.

\begin{acks}
We thank Prof. Sudeshna Sarkar, Prof. Alok Kanti Deb and Prof. Dilip Kumar Pratihar for their constant guidance and support as co-Principal Investigators of Kharagpur RoboSoccer Students' Group (KRSSG). We also thank all the alumni and present members of KRSSG especially Kumar Abhinav, Dhananjay Yadav and Madhav Mantri for addressing game play, mechanical and embedded challenges in robot soccer. This work is supported by Sponsored Research and Industrial Consultancy (SRIC), Indian Institute of Technology, Kharagpur.
\end{acks}


\bibliographystyle{ACM-Reference-Format}
\bibliography{sigproc} 


\begin{thebibliography}{00}


\ifx \showCODEN    \undefined \def \showCODEN     #1{\unskip}     \fi
\ifx \showDOI      \undefined \def \showDOI       #1{{\tt DOI:}\penalty0{#1}\ }
  \fi
\ifx \showISBNx    \undefined \def \showISBNx     #1{\unskip}     \fi
\ifx \showISBNxiii \undefined \def \showISBNxiii  #1{\unskip}     \fi
\ifx \showISSN     \undefined \def \showISSN      #1{\unskip}     \fi
\ifx \showLCCN     \undefined \def \showLCCN      #1{\unskip}     \fi
\ifx \shownote     \undefined \def \shownote      #1{#1}          \fi
\ifx \showarticletitle \undefined \def \showarticletitle #1{#1}   \fi
\ifx \showURL      \undefined \def \showURL       #1{#1}          \fi
\providecommand\bibfield[2]{#2}
\providecommand\bibinfo[2]{#2}
\providecommand\natexlab[1]{#1}

\bibitem[\protect\citeauthoryear{Bardenet, Brendel, K{\'e}gl, and
  Sebag}{Bardenet et~al\mbox{.}}{2013}]%
        {references:bardenet2013collaborative}
\bibfield{author}{\bibinfo{person}{R{\'e}mi Bardenet},
  \bibinfo{person}{M{\'a}ty{\'a}s Brendel}, \bibinfo{person}{Bal{\'a}zs
  K{\'e}gl}, {and} \bibinfo{person}{Michele Sebag}.}
  \bibinfo{year}{2013}\natexlab{}.
\newblock \showarticletitle{Collaborative hyperparameter tuning.}. In
  \bibinfo{booktitle}{{\em ICML (2)}}. \bibinfo{pages}{199--207}.
\newblock


\bibitem[\protect\citeauthoryear{Bergstra, Bardenet, Bengio, and
  K{\'e}gl}{Bergstra et~al\mbox{.}}{2011}]%
        {references:bergstra2011algorithms}
\bibfield{author}{\bibinfo{person}{James~S Bergstra}, \bibinfo{person}{R{\'e}mi
  Bardenet}, \bibinfo{person}{Yoshua Bengio}, {and} \bibinfo{person}{Bal{\'a}zs
  K{\'e}gl}.} \bibinfo{year}{2011}\natexlab{}.
\newblock \showarticletitle{Algorithms for hyper-parameter optimization}. In
  \bibinfo{booktitle}{{\em Advances in Neural Information Processing Systems}}.
  \bibinfo{pages}{2546--2554}.
\newblock


\bibitem[\protect\citeauthoryear{Borenstein and Koren}{Borenstein and
  Koren}{1991}]%
        {borenstein1991vector}
\bibfield{author}{\bibinfo{person}{Johann Borenstein} {and}
  \bibinfo{person}{Yoram Koren}.} \bibinfo{year}{1991}\natexlab{}.
\newblock \showarticletitle{The vector field histogram-fast obstacle avoidance
  for mobile robots}.
\newblock \bibinfo{journal}{{\em IEEE Transactions on Robotics and
  Automation\/}} \bibinfo{volume}{{7}, 3} (\bibinfo{year}{1991}),
  \bibinfo{pages}{278--288}.
\newblock


\bibitem[\protect\citeauthoryear{Brendel and Schoenauer}{Brendel and
  Schoenauer}{2011}]%
        {references:brendel2011instance}
\bibfield{author}{\bibinfo{person}{M{\'a}ty{\'a}s Brendel} {and}
  \bibinfo{person}{Marc Schoenauer}.} \bibinfo{year}{2011}\natexlab{}.
\newblock \showarticletitle{Instance-Based Parameter Tuning and Learning for
  Evolutionary AI Planning}. In \bibinfo{booktitle}{{\em PAL 2011 3rd Workshop
  on Planning and Learning}}. \bibinfo{pages}{5}.
\newblock


\bibitem[\protect\citeauthoryear{Brock and Khatib}{Brock and Khatib}{1999}]%
        {references:brock1999high}
\bibfield{author}{\bibinfo{person}{Oliver Brock} {and} \bibinfo{person}{Oussama
  Khatib}.} \bibinfo{year}{1999}\natexlab{}.
\newblock \showarticletitle{High-speed navigation using the global dynamic
  window approach}. In \bibinfo{booktitle}{{\em Robotics and Automation, 1999.
  Proceedings. 1999 IEEE International Conference on}},
  \bibinfo{volume}{Vol.~1}. IEEE, \bibinfo{pages}{341--346}.
\newblock


\bibitem[\protect\citeauthoryear{Calandra, Seyfarth, Peters, and
  Deisenroth}{Calandra et~al\mbox{.}}{2016}]%
        {references:calandra2016bayesian}
\bibfield{author}{\bibinfo{person}{Roberto Calandra},
  \bibinfo{person}{Andr{\'e} Seyfarth}, \bibinfo{person}{Jan Peters}, {and}
  \bibinfo{person}{Marc~Peter Deisenroth}.} \bibinfo{year}{2016}\natexlab{}.
\newblock \showarticletitle{Bayesian optimization for learning gaits under
  uncertainty}.
\newblock \bibinfo{journal}{{\em Annals of Mathematics and Artificial
  Intelligence\/}} \bibinfo{volume}{{76}, 1-2} (\bibinfo{year}{2016}),
  \bibinfo{pages}{5--23}.
\newblock


\bibitem[\protect\citeauthoryear{Cover and Hart}{Cover and Hart}{1967}]%
        {references:cover1967nearest}
\bibfield{author}{\bibinfo{person}{Thomas Cover} {and} \bibinfo{person}{Peter
  Hart}.} \bibinfo{year}{1967}\natexlab{}.
\newblock \showarticletitle{Nearest neighbor pattern classification}.
\newblock \bibinfo{journal}{{\em IEEE transactions on information theory\/}}
  \bibinfo{volume}{{13}, 1} (\bibinfo{year}{1967}), \bibinfo{pages}{21--27}.
\newblock


\bibitem[\protect\citeauthoryear{Cox and John}{Cox and John}{1992}]%
        {references:cox1992statistical}
\bibfield{author}{\bibinfo{person}{Dennis~D Cox} {and} \bibinfo{person}{Susan
  John}.} \bibinfo{year}{1992}\natexlab{}.
\newblock \showarticletitle{A statistical method for global optimization}. In
  \bibinfo{booktitle}{{\em Systems, Man and Cybernetics, 1992., IEEE
  International Conference on}}. IEEE, \bibinfo{pages}{1241--1246}.
\newblock


\bibitem[\protect\citeauthoryear{De~Luca, Oriolo, and Vendittelli}{De~Luca
  et~al\mbox{.}}{2001}]%
        {references:de2001control}
\bibfield{author}{\bibinfo{person}{Alessandro De~Luca},
  \bibinfo{person}{Giuseppe Oriolo}, {and} \bibinfo{person}{Marilena
  Vendittelli}.} \bibinfo{year}{2001}\natexlab{}.
\newblock \showarticletitle{Control of wheeled mobile robots: An experimental
  overview}.
\newblock In \bibinfo{booktitle}{{\em Ramsete}}. Springer,
  \bibinfo{pages}{181--226}.
\newblock


\bibitem[\protect\citeauthoryear{Farin}{Farin}{2014}]%
        {references:farin2014curves}
\bibfield{author}{\bibinfo{person}{Gerald Farin}.}
  \bibinfo{year}{2014}\natexlab{}.
\newblock \bibinfo{booktitle}{{\em Curves and surfaces for computer-aided
  geometric design: a practical guide}}.
\newblock Elsevier.
\newblock


\bibitem[\protect\citeauthoryear{Hansen, M{\"u}ller, and Koumoutsakos}{Hansen
  et~al\mbox{.}}{2003}]%
        {references:hansen2003reducing}
\bibfield{author}{\bibinfo{person}{Nikolaus Hansen}, \bibinfo{person}{Sibylle~D
  M{\"u}ller}, {and} \bibinfo{person}{Petros Koumoutsakos}.}
  \bibinfo{year}{2003}\natexlab{}.
\newblock \showarticletitle{Reducing the time complexity of the derandomized
  evolution strategy with covariance matrix adaptation (CMA-ES)}.
\newblock \bibinfo{journal}{{\em Evolutionary computation\/}}
  \bibinfo{volume}{{11}, 1} (\bibinfo{year}{2003}), \bibinfo{pages}{1--18}.
\newblock


\bibitem[\protect\citeauthoryear{Hoos and Leyton-Brown}{Hoos and
  Leyton-Brown}{2014}]%
        {references:hoos2014efficient}
\bibfield{author}{\bibinfo{person}{Holger Hoos} {and} \bibinfo{person}{Kevin
  Leyton-Brown}.} \bibinfo{year}{2014}\natexlab{}.
\newblock \showarticletitle{An efficient approach for assessing hyperparameter
  importance}.
\newblock  (\bibinfo{year}{2014}).
\newblock


\bibitem[\protect\citeauthoryear{Jolly, Kumar, and Vijayakumar}{Jolly
  et~al\mbox{.}}{2009}]%
        {references:jolly2009bezier}
\bibfield{author}{\bibinfo{person}{KG Jolly}, \bibinfo{person}{R~Sreerama
  Kumar}, {and} \bibinfo{person}{R Vijayakumar}.}
  \bibinfo{year}{2009}\natexlab{}.
\newblock \showarticletitle{A Bezier curve based path planning in a multi-agent
  robot soccer system without violating the acceleration limits}.
\newblock \bibinfo{journal}{{\em Robotics and Autonomous Systems\/}}
  \bibinfo{volume}{{57}, 1} (\bibinfo{year}{2009}), \bibinfo{pages}{23--33}.
\newblock


\bibitem[\protect\citeauthoryear{Khatib}{Khatib}{1986}]%
        {khatib1986real}
\bibfield{author}{\bibinfo{person}{Oussama Khatib}.}
  \bibinfo{year}{1986}\natexlab{}.
\newblock \showarticletitle{Real-time obstacle avoidance for manipulators and
  mobile robots}.
\newblock In \bibinfo{booktitle}{{\em Autonomous robot vehicles}}. Springer,
  \bibinfo{pages}{396--404}.
\newblock


\bibitem[\protect\citeauthoryear{Klan{\v{c}}ar and {\v{S}}krjanc}{Klan{\v{c}}ar
  and {\v{S}}krjanc}{2007}]%
        {references:klanvcar2007tracking}
\bibfield{author}{\bibinfo{person}{Gregor Klan{\v{c}}ar} {and}
  \bibinfo{person}{Igor {\v{S}}krjanc}.} \bibinfo{year}{2007}\natexlab{}.
\newblock \showarticletitle{Tracking-error model-based predictive control for
  mobile robots in real time}.
\newblock \bibinfo{journal}{{\em Robotics and Autonomous Systems\/}}
  \bibinfo{volume}{{55}, 6} (\bibinfo{year}{2007}), \bibinfo{pages}{460--469}.
\newblock


\bibitem[\protect\citeauthoryear{Kunigahalli and Russell}{Kunigahalli and
  Russell}{1994}]%
        {kunigahalli1994visibility}
\bibfield{author}{\bibinfo{person}{R Kunigahalli} {and} \bibinfo{person}{JS
  Russell}.} \bibinfo{year}{1994}\natexlab{}.
\newblock \showarticletitle{Visibility graph approach to detailed path planning
  in cnc concrete placement}.
\newblock \bibinfo{journal}{{\em Automation and robotics in construction XI\/}}
  (\bibinfo{year}{1994}).
\newblock


\bibitem[\protect\citeauthoryear{Lau, Sprunk, and Burgard}{Lau
  et~al\mbox{.}}{2009}]%
        {references:lau2009kinodynamic}
\bibfield{author}{\bibinfo{person}{Boris Lau}, \bibinfo{person}{Christoph
  Sprunk}, {and} \bibinfo{person}{Wolfram Burgard}.}
  \bibinfo{year}{2009}\natexlab{}.
\newblock \showarticletitle{Kinodynamic motion planning for mobile robots using
  splines}. In \bibinfo{booktitle}{{\em 2009 IEEE/RSJ International Conference
  on Intelligent Robots and Systems}}. IEEE, \bibinfo{pages}{2427--2433}.
\newblock


\bibitem[\protect\citeauthoryear{Lepeti{\v{c}}, Klan{\v{c}}ar, {\v{S}}krjanc,
  Matko, and Poto{\v{c}}nik}{Lepeti{\v{c}} et~al\mbox{.}}{2003}]%
        {references:lepetivc2003time}
\bibfield{author}{\bibinfo{person}{Marko Lepeti{\v{c}}},
  \bibinfo{person}{Gregor Klan{\v{c}}ar}, \bibinfo{person}{Igor {\v{S}}krjanc},
  \bibinfo{person}{Drago Matko}, {and} \bibinfo{person}{Bo{\v{s}}tjan
  Poto{\v{c}}nik}.} \bibinfo{year}{2003}\natexlab{}.
\newblock \showarticletitle{Time optimal path planning considering acceleration
  limits}.
\newblock \bibinfo{journal}{{\em Robotics and Autonomous Systems\/}}
  \bibinfo{volume}{{45}, 3} (\bibinfo{year}{2003}), \bibinfo{pages}{199--210}.
\newblock


\bibitem[\protect\citeauthoryear{Liaw and Wiener}{Liaw and Wiener}{2002}]%
        {liaw2002classification}
\bibfield{author}{\bibinfo{person}{Andy Liaw} {and} \bibinfo{person}{Matthew
  Wiener}.} \bibinfo{year}{2002}\natexlab{}.
\newblock \showarticletitle{Classification and regression by randomForest}.
\newblock \bibinfo{journal}{{\em R news\/}} \bibinfo{volume}{{2}, 3}
  (\bibinfo{year}{2002}), \bibinfo{pages}{18--22}.
\newblock


\bibitem[\protect\citeauthoryear{Lizotte, Wang, Bowling, and
  Schuurmans}{Lizotte et~al\mbox{.}}{2007}]%
        {references:lizotte2007automatic}
\bibfield{author}{\bibinfo{person}{Daniel~J Lizotte}, \bibinfo{person}{Tao
  Wang}, \bibinfo{person}{Michael~H Bowling}, {and} \bibinfo{person}{Dale
  Schuurmans}.} \bibinfo{year}{2007}\natexlab{}.
\newblock \showarticletitle{Automatic Gait Optimization with Gaussian Process
  Regression.}. In \bibinfo{booktitle}{{\em IJCAI}}, \bibinfo{volume}{Vol.~7}.
  \bibinfo{pages}{944--949}.
\newblock


\bibitem[\protect\citeauthoryear{Mahkovic and Slivnik}{Mahkovic and
  Slivnik}{1997}]%
        {mahkovic1997smooth}
\bibfield{author}{\bibinfo{person}{Rajko Mahkovic} {and}
  \bibinfo{person}{Toma{\v{z}} Slivnik}.} \bibinfo{year}{1997}\natexlab{}.
\newblock \showarticletitle{Smooth Path Planning on the Basis of Reduced
  Visibility Graph}.
\newblock \bibinfo{journal}{{\em MODELLING IDENTIFICATION AND CONTROL\/}}
  (\bibinfo{year}{1997}), \bibinfo{pages}{267--270}.
\newblock


\bibitem[\protect\citeauthoryear{Martinez-Cantin}{Martinez-Cantin}{2014}]%
        {references:martinez2014bayesopt}
\bibfield{author}{\bibinfo{person}{Ruben Martinez-Cantin}.}
  \bibinfo{year}{2014}\natexlab{}.
\newblock \showarticletitle{BayesOpt: a Bayesian optimization library for
  nonlinear optimization, experimental design and bandits.}
\newblock \bibinfo{journal}{{\em Journal of Machine Learning Research\/}}
  \bibinfo{volume}{{15}, 1} (\bibinfo{year}{2014}),
  \bibinfo{pages}{3735--3739}.
\newblock


\bibitem[\protect\citeauthoryear{Martinez-Cantin, de~Freitas, Brochu,
  Castellanos, and Doucet}{Martinez-Cantin et~al\mbox{.}}{2009}]%
        {references:martinez2009bayesian}
\bibfield{author}{\bibinfo{person}{Ruben Martinez-Cantin},
  \bibinfo{person}{Nando de Freitas}, \bibinfo{person}{Eric Brochu},
  \bibinfo{person}{Jos{\'e} Castellanos}, {and} \bibinfo{person}{Arnaud
  Doucet}.} \bibinfo{year}{2009}\natexlab{}.
\newblock \showarticletitle{A Bayesian exploration-exploitation approach for
  optimal online sensing and planning with a visually guided mobile robot}.
\newblock \bibinfo{journal}{{\em Autonomous Robots\/}} \bibinfo{volume}{{27},
  2} (\bibinfo{year}{2009}), \bibinfo{pages}{93--103}.
\newblock


\bibitem[\protect\citeauthoryear{Martinez-Cantin, de~Freitas, Doucet, and
  Castellanos}{Martinez-Cantin et~al\mbox{.}}{2007}]%
        {references:martinez2007active}
\bibfield{author}{\bibinfo{person}{Ruben Martinez-Cantin},
  \bibinfo{person}{Nando de Freitas}, \bibinfo{person}{Arnaud Doucet}, {and}
  \bibinfo{person}{Jos{\'e}~A Castellanos}.} \bibinfo{year}{2007}\natexlab{}.
\newblock \showarticletitle{Active Policy Learning for Robot Planning and
  Exploration under Uncertainty.}. In \bibinfo{booktitle}{{\em Robotics:
  Science and Systems}}. \bibinfo{pages}{321--328}.
\newblock


\bibitem[\protect\citeauthoryear{Meng and Picton}{Meng and Picton}{1992}]%
        {meng1992neural}
\bibfield{author}{\bibinfo{person}{H Meng} {and} \bibinfo{person}{PD Picton}.}
  \bibinfo{year}{1992}\natexlab{}.
\newblock \showarticletitle{A neural network for collision-free path planning}.
\newblock \bibinfo{journal}{{\em Artificial neural networks\/}}
  \bibinfo{volume}{{2}, 1} (\bibinfo{year}{1992}), \bibinfo{pages}{591--4}.
\newblock


\bibitem[\protect\citeauthoryear{Nguyen, Chen, and Ng}{Nguyen
  et~al\mbox{.}}{2007}]%
        {references:nguyen2007planning}
\bibfield{author}{\bibinfo{person}{Kim~Doang Nguyen}, \bibinfo{person}{I-Ming
  Chen}, {and} \bibinfo{person}{Teck-Chew Ng}.}
  \bibinfo{year}{2007}\natexlab{}.
\newblock \showarticletitle{Planning algorithms for s-curve trajectories}. In
  \bibinfo{booktitle}{{\em 2007 IEEE/ASME international conference on advanced
  intelligent mechatronics}}. IEEE, \bibinfo{pages}{1--6}.
\newblock


\bibitem[\protect\citeauthoryear{Nguyen, Ng, and Chen}{Nguyen
  et~al\mbox{.}}{2008}]%
        {references:nguyen2008algorithms}
\bibfield{author}{\bibinfo{person}{Kim~Doang Nguyen},
  \bibinfo{person}{Teck-Chew Ng}, {and} \bibinfo{person}{I-Ming Chen}.}
  \bibinfo{year}{2008}\natexlab{}.
\newblock \showarticletitle{On algorithms for planning S-curve motion
  profiles}.
\newblock \bibinfo{journal}{{\em International Journal of Advanced Robotic
  Systems\/}} \bibinfo{volume}{{5}, 1} (\bibinfo{year}{2008}),
  \bibinfo{pages}{99--106}.
\newblock


\bibitem[\protect\citeauthoryear{Rasmussen}{Rasmussen}{2006}]%
        {references:rasmussen2006gaussian}
\bibfield{author}{\bibinfo{person}{Carl~Edward Rasmussen}.}
  \bibinfo{year}{2006}\natexlab{}.
\newblock \showarticletitle{Gaussian processes for machine learning}.
\newblock  (\bibinfo{year}{2006}).
\newblock


\bibitem[\protect\citeauthoryear{Ribeiro}{Ribeiro}{2004}]%
        {references:ribeiro2004kalman}
\bibfield{author}{\bibinfo{person}{Maria~Isabel Ribeiro}.}
  \bibinfo{year}{2004}\natexlab{}.
\newblock \showarticletitle{Kalman and extended kalman filters: Concept,
  derivation and properties}.
\newblock \bibinfo{journal}{{\em Institute for Systems and Robotics\/}}
  \bibinfo{volume}{43} (\bibinfo{year}{2004}).
\newblock


\bibitem[\protect\citeauthoryear{Rosman, Hawasly, and Ramamoorthy}{Rosman
  et~al\mbox{.}}{2016}]%
        {references:rosman2016bayesian}
\bibfield{author}{\bibinfo{person}{Benjamin Rosman}, \bibinfo{person}{Majd
  Hawasly}, {and} \bibinfo{person}{Subramanian Ramamoorthy}.}
  \bibinfo{year}{2016}\natexlab{}.
\newblock \showarticletitle{Bayesian policy reuse}.
\newblock \bibinfo{journal}{{\em Machine Learning\/}} \bibinfo{volume}{{104},
  1} (\bibinfo{year}{2016}), \bibinfo{pages}{99--127}.
\newblock


\bibitem[\protect\citeauthoryear{Sahraei, Manzuri, Razvan, Tajfard, and
  Khoshbakht}{Sahraei et~al\mbox{.}}{2007}]%
        {references:sahraei2007real}
\bibfield{author}{\bibinfo{person}{Alireza Sahraei},
  \bibinfo{person}{Mohammad~Taghi Manzuri}, \bibinfo{person}{Mohammad~Reza
  Razvan}, \bibinfo{person}{Masoud Tajfard}, {and} \bibinfo{person}{Saman
  Khoshbakht}.} \bibinfo{year}{2007}\natexlab{}.
\newblock \showarticletitle{Real-time trajectory generation for mobile robots}.
  In \bibinfo{booktitle}{{\em Congress of the Italian Association for
  Artificial Intelligence}}. Springer, \bibinfo{pages}{459--470}.
\newblock


\bibitem[\protect\citeauthoryear{Saska, Kulich, Klancar, and Faigl}{Saska
  et~al\mbox{.}}{2006a}]%
        {saska2006transformed}
\bibfield{author}{\bibinfo{person}{M Saska}, \bibinfo{person}{M Kulich},
  \bibinfo{person}{G Klancar}, {and} \bibinfo{person}{J Faigl}.}
  \bibinfo{year}{2006}\natexlab{a}.
\newblock \showarticletitle{Transformed net-collision avoidance algorithm for
  robotic soccer}. In \bibinfo{booktitle}{{\em Proceedings 5th MATHMOD
  Vienna-5th Vienna Symposium on Mathematical Modelling}}.
\newblock


\bibitem[\protect\citeauthoryear{Saska, Macas, Preucil, and Lhotska}{Saska
  et~al\mbox{.}}{2006b}]%
        {saska2006robot}
\bibfield{author}{\bibinfo{person}{Martin Saska}, \bibinfo{person}{Martin
  Macas}, \bibinfo{person}{Libor Preucil}, {and} \bibinfo{person}{Lenka
  Lhotska}.} \bibinfo{year}{2006}\natexlab{b}.
\newblock \showarticletitle{Robot path planning using particle swarm
  optimization of Ferguson splines}. In \bibinfo{booktitle}{{\em Emerging
  Technologies and Factory Automation, 2006. ETFA'06. IEEE Conference on}}.
  IEEE, \bibinfo{pages}{833--839}.
\newblock


\bibitem[\protect\citeauthoryear{Schonlau, Welch, and Jones}{Schonlau
  et~al\mbox{.}}{1998}]%
        {references:schonlau1998global}
\bibfield{author}{\bibinfo{person}{Matthias Schonlau},
  \bibinfo{person}{William~J Welch}, {and} \bibinfo{person}{Donald~R Jones}.}
  \bibinfo{year}{1998}\natexlab{}.
\newblock \showarticletitle{Global versus local search in constrained
  optimization of computer models}.
\newblock \bibinfo{journal}{{\em Lecture Notes-Monograph Series\/}}
  (\bibinfo{year}{1998}), \bibinfo{pages}{11--25}.
\newblock


\bibitem[\protect\citeauthoryear{Seder and Petrovic}{Seder and
  Petrovic}{2007}]%
        {references:seder2007dynamic}
\bibfield{author}{\bibinfo{person}{Marija Seder} {and} \bibinfo{person}{Ivan
  Petrovic}.} \bibinfo{year}{2007}\natexlab{}.
\newblock \showarticletitle{Dynamic window based approach to mobile robot
  motion control in the presence of moving obstacles}. In
  \bibinfo{booktitle}{{\em Proceedings 2007 IEEE International Conference on
  Robotics and Automation}}. IEEE, \bibinfo{pages}{1986--1991}.
\newblock


\bibitem[\protect\citeauthoryear{Shiller and Gwo}{Shiller and Gwo}{1991}]%
        {references:shiller1991dynamic}
\bibfield{author}{\bibinfo{person}{Zvi Shiller} {and} \bibinfo{person}{Y-R
  Gwo}.} \bibinfo{year}{1991}\natexlab{}.
\newblock \showarticletitle{Dynamic motion planning of autonomous vehicles}.
\newblock \bibinfo{journal}{{\em IEEE Transactions on Robotics and
  Automation\/}} \bibinfo{volume}{{7}, 2} (\bibinfo{year}{1991}),
  \bibinfo{pages}{241--249}.
\newblock


\bibitem[\protect\citeauthoryear{Snoek, Larochelle, and Adams}{Snoek
  et~al\mbox{.}}{2012}]%
        {references:snoek2012practical}
\bibfield{author}{\bibinfo{person}{Jasper Snoek}, \bibinfo{person}{Hugo
  Larochelle}, {and} \bibinfo{person}{Ryan~P Adams}.}
  \bibinfo{year}{2012}\natexlab{}.
\newblock \showarticletitle{Practical bayesian optimization of machine learning
  algorithms}. In \bibinfo{booktitle}{{\em Advances in neural information
  processing systems}}. \bibinfo{pages}{2951--2959}.
\newblock


\bibitem[\protect\citeauthoryear{Souza, Marchant, Ott, Wolf, and Ramos}{Souza
  et~al\mbox{.}}{2014}]%
        {references:souza2014bayesian}
\bibfield{author}{\bibinfo{person}{Jefferson~R Souza}, \bibinfo{person}{Roman
  Marchant}, \bibinfo{person}{Lionel Ott}, \bibinfo{person}{Denis~F Wolf},
  {and} \bibinfo{person}{Fabio Ramos}.} \bibinfo{year}{2014}\natexlab{}.
\newblock \showarticletitle{Bayesian optimisation for active perception and
  smooth navigation}. In \bibinfo{booktitle}{{\em Robotics and Automation
  (ICRA), 2014 IEEE International Conference on}}. IEEE,
  \bibinfo{pages}{4081--4087}.
\newblock


\bibitem[\protect\citeauthoryear{Sprunk}{Sprunk}{2008}]%
        {references:sprunk2008planning}
\bibfield{author}{\bibinfo{person}{Christoph Sprunk}.}
  \bibinfo{year}{2008}\natexlab{}.
\newblock \showarticletitle{Planning motion trajectories for mobile robots
  using splines}.
\newblock \bibinfo{journal}{{\em University of Freiburg\/}}
  (\bibinfo{year}{2008}).
\newblock


\bibitem[\protect\citeauthoryear{Stein}{Stein}{1987}]%
        {references:stein1987large}
\bibfield{author}{\bibinfo{person}{Michael Stein}.}
  \bibinfo{year}{1987}\natexlab{}.
\newblock \showarticletitle{Large sample properties of simulations using Latin
  hypercube sampling}.
\newblock \bibinfo{journal}{{\em Technometrics\/}} \bibinfo{volume}{{29}, 2}
  (\bibinfo{year}{1987}), \bibinfo{pages}{143--151}.
\newblock


\bibitem[\protect\citeauthoryear{Visioli}{Visioli}{2001}]%
        {references:visioli2001tuning}
\bibfield{author}{\bibinfo{person}{Antonio Visioli}.}
  \bibinfo{year}{2001}\natexlab{}.
\newblock \showarticletitle{Tuning of PID controllers with fuzzy logic}.
\newblock \bibinfo{journal}{{\em IEE Proceedings-Control Theory and
  Applications\/}} \bibinfo{volume}{{148}, 1} (\bibinfo{year}{2001}),
  \bibinfo{pages}{1--8}.
\newblock


\bibitem[\protect\citeauthoryear{Walter and Fournier}{Walter and
  Fournier}{1996}]%
        {references:walter1996approximate}
\bibfield{author}{\bibinfo{person}{Marcelo Walter} {and} \bibinfo{person}{Alain
  Fournier}.} \bibinfo{year}{1996}\natexlab{}.
\newblock \showarticletitle{Approximate arc length parameterization}. In
  \bibinfo{booktitle}{{\em Proceedings of the 9th Brazilian symposium on
  computer graphics and image processing}}. Citeseer,
  \bibinfo{pages}{143--150}.
\newblock


\bibitem[\protect\citeauthoryear{Wang, Kearney, and Atkinson}{Wang
  et~al\mbox{.}}{2002}]%
        {references:wang2002arc}
\bibfield{author}{\bibinfo{person}{Hongling Wang}, \bibinfo{person}{Joseph
  Kearney}, {and} \bibinfo{person}{Kendall Atkinson}.}
  \bibinfo{year}{2002}\natexlab{}.
\newblock \showarticletitle{Arc-length parameterized spline curves for
  real-time simulation}. In \bibinfo{booktitle}{{\em 5th international
  conference on Curves and Surfaces}}.
\newblock


\bibitem[\protect\citeauthoryear{Wang, Liu, Deng, and Xu}{Wang
  et~al\mbox{.}}{2006}]%
        {wang2006obstacle}
\bibfield{author}{\bibinfo{person}{Li Wang}, \bibinfo{person}{Yushu Liu},
  \bibinfo{person}{Hongbin Deng}, {and} \bibinfo{person}{Yuanqing Xu}.}
  \bibinfo{year}{2006}\natexlab{}.
\newblock \showarticletitle{Obstacle-avoidance path planning for soccer robots
  using particle swarm optimization}. In \bibinfo{booktitle}{{\em Robotics and
  Biomimetics, 2006. ROBIO'06. IEEE International Conference on}}. IEEE,
  \bibinfo{pages}{1233--1238}.
\newblock


\end{thebibliography}

\end{document}